\documentclass[letterpaper, 10pt, conference]{ieeeconf}  
\IEEEoverridecommandlockouts    

\usepackage{multirow}
\usepackage{lipsum}
\usepackage{amsmath}
\usepackage{amssymb}
\usepackage[ruled,vlined]{algorithm2e}
\usepackage{graphicx}
\graphicspath{{./Figures/}}
\usepackage{booktabs}
\usepackage[hidelinks]{hyperref}
\usepackage{fancyhdr}

\title{\LARGE \bf Beam-Wise Statistical Background Subtraction for Static Roadside LiDAR: A Cross-Sensor Benchmark Study}

\author{Alexander Baumann$^{1}$, Marcel Vosshans$^{1}$, Thao Dang$^{1}$
\thanks{$^{1}$The authors are with the Institute for Intelligent Systems which is part of the Faculty of Computer Science and Engineering, University of Applied Sciences Esslingen, Germany
        {\tt\small \{alexander.baumann, marcel.vosshans, thao.dang\}@hs-esslingen.de}}%
}

\begin{document}
\maketitle
\thispagestyle{fancy}
\fancyhf{}
\renewcommand{\headrulewidth}{0pt}
\renewcommand{\footrulewidth}{0pt}
\fancyfoot[C]{\footnotesize
  \begin{minipage}{0.95\textwidth}
    \centering
    \textcopyright~2026 IEEE.
    Personal use of this material is permitted. Permission from IEEE must be obtained for all other uses, in any current or future media, including reprinting/republishing this material for advertising or promotional purposes, creating new collective works, for resale or redistribution to
    servers or lists, or reuse of any copyrighted component of this work in other works.\\
    Accepted for publication in the proceedings of the 29th IEEE International Conference on Intelligent Transportation Systems (ITSC 2026), Naples, Italy, September 15--18, 2026.
  \end{minipage}
}
\thispagestyle{fancy}
\pagestyle{empty}


\begin{abstract}
Background subtraction is a key preprocessing step for infrastructure-based LiDAR perception, enabling efficient isolation of dynamic traffic participants without semantic annotations. However, systematic cross-sensor evaluations and reproducible studies for static roadside LiDAR are missing.

This paper presents a comparative benchmark of beam-wise statistical background subtraction for statically mounted LiDAR sensors. We formulate background estimation as a per-beam temporal modeling problem and investigate complementary statistical strategies that capture dominant as well as multi-modal background structures, combined with spatial filtering in the angular and 3D domain.
To enable reproducible evaluation, we introduce \textit{HighwayScene}, a new multi-LiDAR dataset recorded in a static roadside setup, and extend the public \textit{CoopScenes} dataset with static/dynamic point-wise annotations.
Across multiple scenes and heterogeneous sensing technologies, we demonstrate that beam-wise statistical modeling provides a robust and transferable solution. Combining lightweight per-beam models with spatial consistency filtering substantially improves precision while maintaining high recall and real-time capability. All datasets, annotations, and implementations are publicly released.
\end{abstract}

\section{Introduction}
\label{sec:introduction}

Infrastructure-based perception is becoming increasingly important in intelligent transportation systems.
Vehicle-to-Infrastructure (V2I) communication is evolving from simple warning systems toward cooperative and automated driving applications \cite{wei_cooperativeperceptionautomated_2025}.
Beyond direct vehicle interaction, smart infrastructure deployed at critical traffic junctions enables improved traffic monitoring, adaptive traffic control, and data-driven urban planning.
Infrastructure-mounted sensors further support structural health monitoring of roads and bridges by providing continuous measurements of traffic density, vehicle spacing, and flow dynamics.

Among the available sensing modalities, LiDAR has emerged as a promising candidate for roadside deployment.
In contrast to camera-based systems, LiDAR is largely invariant to illumination changes and provides reliable measurements under challenging lighting conditions such as glare, shadows, or at night \cite{yurtsever_surveyautonomousdriving_2020}.
Furthermore, camera systems may raise privacy concerns due to the recording of identifiable visual information \cite{sookhak_securityprivacysmart_2019}.
LiDAR, in comparison, captures geometric structures without recording identifiable visual information.
Its direct range measurements additionally enable applications that require precise distance estimation, e.g. monitoring inter-vehicle distances on bridges to prevent structural overload caused by closely spaced heavy trucks.

\begin{figure}[t]
    \centering
    \includegraphics[width=0.95\linewidth]{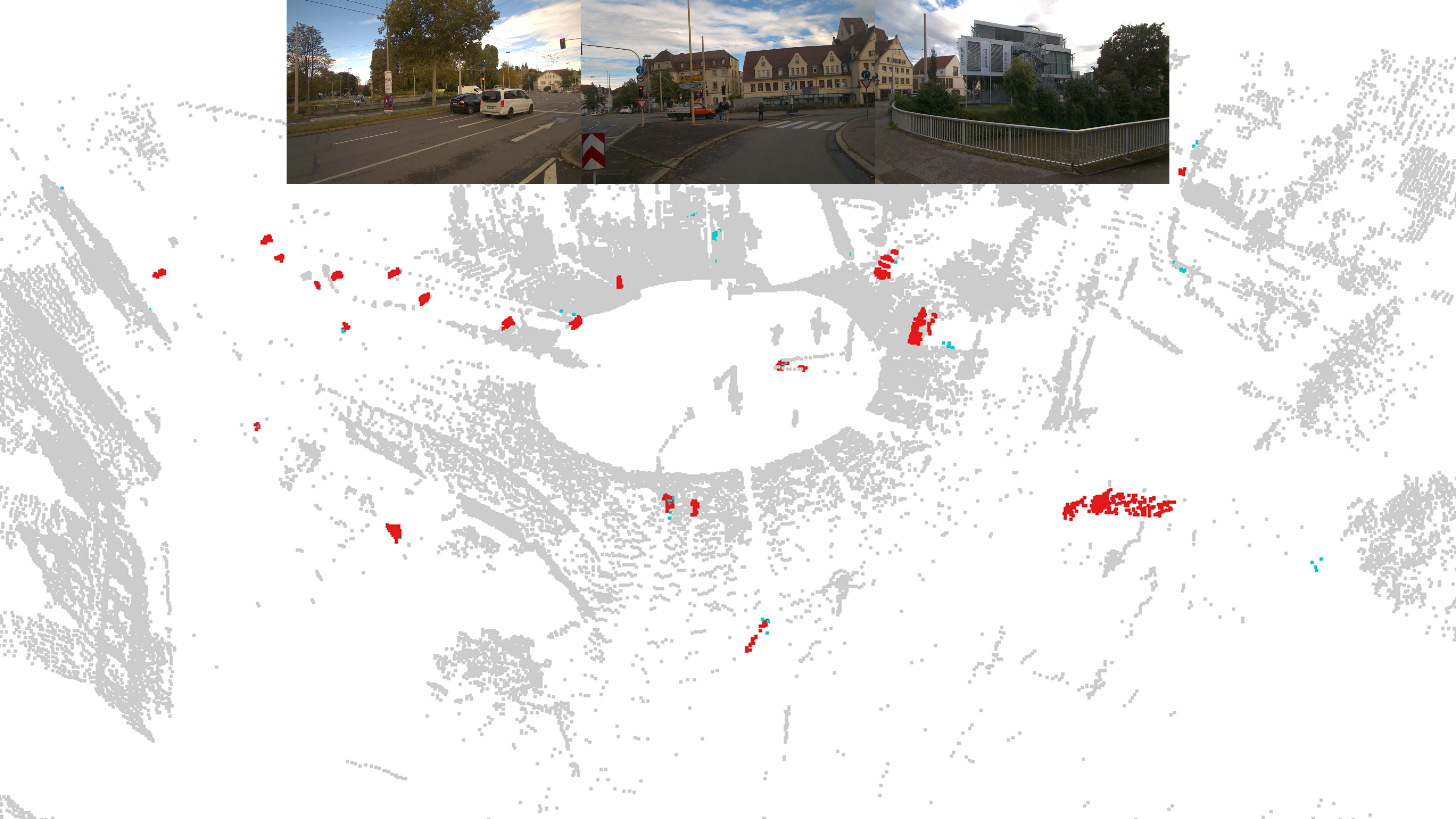}
    \caption{
    Example result of the density-aware coarse-fine triangle algorithm (CFTA) with spatial consistency filtering on the CoopScenes dataset.
    Correctly classified points are shown in light gray (background) and red (foreground), while misclassified points are shown in cyan.
    }
    \label{fig:intro_result}
\end{figure}

In many infrastructure-based perception tasks, the primary objective is the detection and analysis of dynamic objects.
Background subtraction therefore represents a fundamental preprocessing step.
By identifying the static scene structure, background subtraction reduces data volume to the relevant dynamic components and directly isolates moving traffic participants (see Figure \ref{fig:intro_result}).
While object detectors developed for ego-vehicle perception have achieved strong performance, they require costly labeled data and often generalize poorly across different scenes \cite{zhang_stal3dunsuperviseddomain_2024}.
Statically mounted LiDAR sensors leverage the fixed observation geometry to enable scene-specific background modeling without relying on semantic annotations.

Despite increasing interest in roadside LiDAR sensing, robust and reproducible background subtraction approaches remain limited.
Existing methods frequently rely on restrictive assumptions, such as dominant background  \cite{zhang_roadsidelidarvehicle_2022}, predefined regions of interest, or specific traffic conditions \cite{zhang_backgroundfilteringvehicle_2018}.
Publicly available implementations are scarce, and systematic evaluations across heterogeneous sensors and scenes are missing.
The absence of suitable benchmark datasets with static/dynamic ground-truth labels further limits fair comparison and reproducibility.

This work addresses these limitations by conducting a systematic study of beam-wise statistical background modeling for static LiDAR infrastructure setups.
We investigate multiple modeling paradigms under varying traffic densities, sensor technologies, scan patterns, and ground-truth generation strategies, aiming to identify robust and transferable design choices for infrastructure-based LiDAR background subtraction.

\subsection{Contributions}

\begin{enumerate}

\item \textbf{Infrastructure LiDAR Dataset.}
We introduce \emph{HighwayScene}, a multi-LiDAR highway dataset recorded in a static roadside setup using solid-state, rotating Time-of-Flight (ToF), and Frequency-Modulated Continuous-Wave (FMCW) LiDAR sensors.
In addition, we provide static/dynamic point-wise annotations for four scenes of the public CoopScenes dataset.

\item \textbf{Beam-Wise Statistical Background Modeling.}
We formulate background subtraction as a beam-wise statistical estimation problem and investigate complementary modeling strategies, including density-aware histogram thresholding, robust quantile-constrained Gaussian modeling, and stability-driven Gaussian mixture modeling.
We further analyze range-adaptive and angular grid-based spatial consistency filtering.

\item \textbf{Cross-Sensor Benchmark Study.}
We present a systematic evaluation across five scenes from two datasets, covering multiple LiDAR technologies, scan patterns, and traffic conditions.
The study provides a structured comparison of modeling assumptions and spatial consistency strategies for static roadside LiDAR.
Since all resources will be made publicly available upon publication of the paper, the benchmark is fully reproducible for future developments.
    
\end{enumerate}

Public resources released alongside this paper include the HighwayScene dataset, the CoopScenes static/dynamic annotations, and the LiDAR background subtraction pipeline.\footnote{\url{https://highwayscene.github.io/}}

\subsection{Paper Organization}

We begin with a review of related work in Section \ref{sec:relwork}. 
Section \ref{sec:datasets} introduces the employed datasets, including the newly recorded HighwayScene dataset and the extended CoopScenes annotations, together with the corresponding ground-truth generation strategies. 
The proposed beam-wise statistical background modeling approaches and spatial consistency filtering methods are presented in Section \ref{sec:backgroundmodelingmethods}. 
Experimental results and the cross-sensor evaluation are reported in Section \ref{sec:experiments}. 
Finally, Section \ref{sec:conclusion} concludes the paper and outlines directions for future work.
\section{Related Work}
\label{sec:relwork}

\begin{figure*}[t]
    \centering
    \def\svgwidth{0.98\textwidth}
    \input{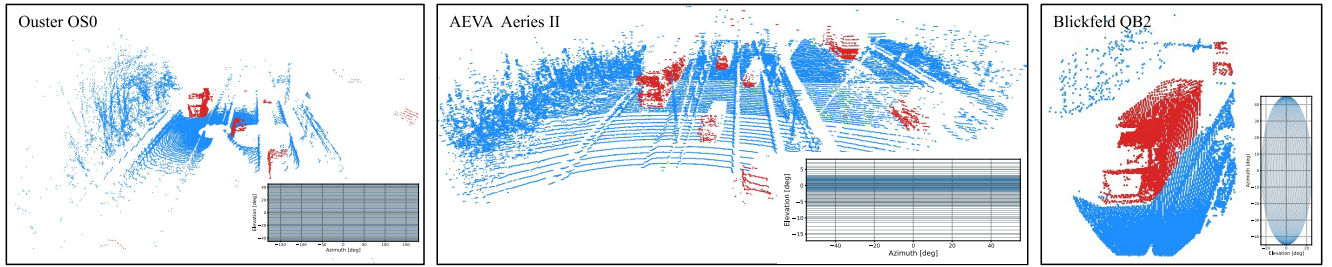}
        \caption{
        Overview of the HighwayScene dataset.
        The figure presents representative labeled frames from three statically mounted LiDAR sensors
        based on rotating, FMCW, and solid-state sensing principles (background: blue, foreground: red).
        Insets depict the corresponding scan patterns, illustrating differences in angular sampling and field-of-view.
        Combined with the use of ROI-based and velocity-based annotation strategies,
        these characteristics create heterogeneous conditions for evaluating background subtraction methods.
        }
    \label{fig:highwayscene_overview}
\end{figure*}

A variety of approaches have been proposed for background subtraction in roadside LiDAR point clouds. 
Early methods relied on the explicit selection of background-only frames. 
Zhang et al. \cite{zhang_backgroundfilteringvehicle_2018} manually selected frames without foreground objects and filtered dynamic points using spatial point association and clustering.
Although effective in controlled scenarios, such approaches depend on the availability of frames that are free of foreground objects, which is difficult to guarantee in real-world traffic scenes.

To avoid manual or explicit background frame selection, several methods model the background statistically over an extended observation period. 
Voxel-based approaches discretize the 3D space and estimate background likelihoods based on point density statistics. 
Wu et al. \cite{wu_automaticbackgroundfiltering_2017} introduced the 3D Density-Statistic Filtering (3D-DSF) method, which identifies background voxels by aggregating point densities over many frames. 
Ansari et al. \cite{ansari_realtimetrafficdata_2025} extend this idea by introducing distance-adaptive thresholds to mitigate the range-dependent sparsity of LiDAR measurements. 
Other works incorporate temporal statistics \cite{lv_rasterbasedbackgroundfiltering_2019} or density-variation analysis \cite{lin_densityvariationbasedbackground_2023} to improve robustness in high-traffic areas. 
Despite their effectiveness, these voxel-based methods are computationally expensive and scale poorly with increasing traffic density, sensor range, and LiDAR resolution.

An alternative class of methods discretizes the LiDAR measurement space into individual rays and models range statistics on a per-beam basis, forming a 2D angular representation. 
Zhang et al. \cite{zhang_automaticbackgroundconstruction_2020} were among the first to exploit this formulation by deriving beam-specific background thresholds under the assumption that the farthest return along each beam corresponds to static background.
This principle was further extended by applying dynamic mode decomposition (DMD) to the temporal LiDAR intensity measurements to separate a temporally stable background from sparse foreground components, followed by a coarse-fine triangle algorithm (CFTA) for beam-wise range threshold estimation \cite{zhang_roadsidelidarvehicle_2022}.

In real-world roadside LiDAR data, however, the assumption that the farthest return consistently represents background does not always hold. 
Multiple static structures may coexist along a single beam, particularly in the presence of vegetation, thin objects, or low reflectivity surfaces. 
To explicitly account for such multimodal range distributions, adaptive Gaussian mixture formulations have been proposed. 
By modeling the range distribution of each beam as a mixture of Gaussian components, multiple background modes can be represented simultaneously, improving robustness in complex urban environments \cite{wang_adaptivepolargridgaussianmixture_2022}.

These beam–range based methods result in background models that are inherently coupled to a specific scan pattern and sensor configuration. 
While this enables high angular resolution, it limits transferability across sensors with different beam layouts.
Iglesias et al. \cite{iglesias_fullyinterpretablestatistical_2025} address this limitation by pooling neighboring beams into a Gaussian polar grid defined in the angular domain. 
By modeling range statistics over angular cells rather than individual beams, the method achieves sensor independence and provides a more statistically stable and interpretable background model.

Beyond statistical models, learning-based dynamic/static separation has been studied in video background subtraction \cite{braham_deepbackgroundsubtraction_2016}, particularly on benchmarks such as CDNet \cite{wang_cdnet2014expanded_2014}, as well as in autonomous-driving LiDAR perception, where dynamic grid maps have been combined with deep detectors for generic dynamic object detection \cite{yan_deepgenericdynamic_2024}.
However, these methods primarily target image-domain change detection or ego-vehicle object-level motion reasoning, whereas this work focuses on unsupervised per-beam background estimation for fixed roadside LiDAR, where sparsity, viewpoint dependency, and scene specificity make direct transfer difficult.

Another practical challenge is the lack of suitable public datasets for evaluating static LiDAR background subtraction methods.
Most publicly available roadside LiDAR datasets focus on object detection or semantic segmentation and provide class-based labels that do not distinguish between static and dynamic points.
In addition, many datasets are limited to a single sensor or scene. 
To the best of our knowledge, CoopScenes \cite{vosshans_coopscenesmultisceneinfrastructure_2025} is currently the only large-scale public dataset that partially fulfills these requirements by providing multiple LiDAR sensors at varying locations.
\section{Datasets}
\label{sec:datasets}

This work addresses background subtraction for statically mounted LiDAR sensors.
Accordingly, any dataset suitable for this task must be recorded from a fixed viewpoint and provide sufficient scene dynamics to reliably distinguish static background from moving foreground.
Beyond this basic requirement, a meaningful benchmark should cover different LiDAR technologies and scan patterns, multiple recording locations with varying traffic densities and scene complexities, and provide ground-truth annotations that allow quantitative evaluation of static and dynamic points.

Collecting infrastructure-mounted LiDAR data across multiple locations and sensor technologies is highly labor-intensive and difficult to scale.
As a result, such a comprehensive dataset could not be realized within the scope of a single data collection effort.
Instead, this work evaluates the proposed methods on two complementary datasets.
The first, \emph{HighwayScene}, is newly recorded and released as part of this work, while the second, CoopScenes, provides additional diversity in terms of locations and scene complexity \cite{vosshans_coopscenesmultisceneinfrastructure_2025}.

\subsection{HighwayScene}

HighwayScene was recorded at a highway construction site with a legally enforced speed limit of 40\,km/h.
The scene exhibits consistently high traffic volume, while vehicles remain free-flowing and do not come to a complete stop.
The dataset emphasizes differences in LiDAR technologies and alternative ground-truth strategies.
An overview of the sensors and their sensing principles is shown in Figure \ref{fig:highwayscene_overview}.

Three statically mounted LiDAR sensors with different sensing principles are used: a Blickfeld QB2 solid-state, an Ouster OS0 rotating, and an AEVA Aeries~II FMCW LiDAR.

For this work, 5,997 frames are recorded at 10 Hz, with the Blickfeld sensor operating at 5 Hz. All sensor streams are temporally synchronized.

Instead of performing manual point-wise static/dynamic annotation, which is costly and error-prone at large scale, the dataset exploits the structured geometry of the highway section for ground-truth generation.

For each lane, a static three-dimensional region of interest (ROI) is manually defined as a rectangular volume aligned with the roadway, extending slightly above the road surface and covering the full drivable width. Under empty-scene conditions, no LiDAR points fall inside these volumes. Since vehicles are constrained by concrete barriers and cannot change lanes, and pedestrians or cyclists cannot access the roadway, any LiDAR returns within a lane volume originate from passing traffic participants.
Points inside the lane volumes are labeled as dynamic, while points outside are considered static background. This geometric labeling enables reliable static/dynamic separation without manual annotation and is applied to the Ouster and Blickfeld sensors.

For the AEVA FMCW LiDAR, ground truth is generated using radial velocities.
A point is labeled as dynamic if its measured velocity exceeds 1 m/s.
This velocity-based labeling does not rely on scene geometry and illustrates the potential of FMCW sensing for automatic annotation.
Its limitations include measurement noise and temporarily stationary dynamic objects.
While such cases were not observed in this dataset after verification, they could be addressed in more general settings using object-level tracking.

Figure \ref{fig:highwayscene_overview} shows representative labeled point clouds
for all three sensors together with their scan patterns.
The visualization highlights differences in angular sampling and sensing principle,
as well as the distinction between geometry-based and velocity-based ground-truth generation.
These factors directly shape the statistical structure of range observations and thereby influence background modeling behavior.

\subsection{CoopScenes}

CoopScenes is a large-scale dataset for cooperative perception, comprising data from an ego vehicle and multiple infrastructure sensors \cite{vosshans_coopscenesmultisceneinfrastructure_2025}. 
It is used in this work due to the availability of static LiDAR recordings captured with an identical setup across multiple locations.

Experiments are conducted using the infrastructure-mounted Ouster OS2 LiDAR. 
From the nine available locations, four urban scenes are selected to ensure a representative diversity in scene complexity and foreground density: Scenes 1 and 4 represent dense and structurally complex environments with high foreground activity, whereas Scenes 2 and 3 are comparatively sparse. 
Background models are trained independently for each scene.

The dataset does not provide point-wise static/dynamic ground-truth annotations. 
Therefore, validation and test labels are manually annotated and released as part of this work.
Across the four selected scenes, 76 frames are annotated, containing 2455 labeled objects, of which 857 are dynamic. 
Annotations are provided at the object level and mapped to point-wise labels for evaluation. 
This annotation strategy provides a third ground-truth alternative and enables evaluation in complex urban environments where simple geometric priors, such as lane-based ROIs, are not applicable.
\section{Methods}
\label{sec:backgroundmodelingmethods}

In this section, we describe the background modeling and filtering strategies evaluated in this work.
All proposed methods operate on the native angular sampling of the LiDAR sensor. 
Given a statically mounted sensor with a known scan pattern, each measurement is associated with fixed elevation and azimuth angles. 
For each beam $b$, corresponding to a unique elevation--azimuth angle pair, a temporal sequence of range measurements
\[
r_{b,t}, \quad t = 1, \ldots, T
\]
is constructed from the training data. 
Based on this sequence, a beam-wise background distribution is estimated.

The only assumptions shared by all methods are:
(i) the scan pattern is known and fixed, and
(ii) the static background is temporally stable.
The beam-wise formulation enables real-time operation even for high-resolution LiDAR sensors with ranges exceeding 150\,m, as all statistical modeling is performed independently per beam.

\subsection{Background Modeling}
We evaluate four background modeling approaches: three beam-wise methods proposed in this work and one established image-based baseline \cite{zivkovic_efficientadaptivedensity_2006}. The original CFTA formulation \cite{zhang_roadsidelidarvehicle_2022} serves as an additional reference baseline.

\subsubsection{Density-Aware CFTA}

\begin{figure}[t]
    \centering
    \def\svgwidth{0.98\linewidth}
    \input{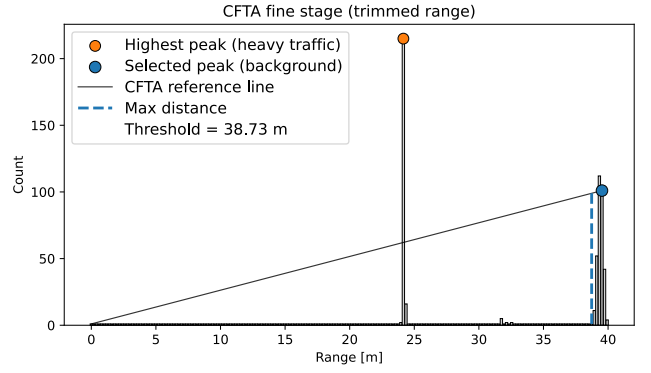}
        \caption{
        Example of a beam-wise range histogram with two competing peaks: a dominant near-range peak caused by dynamic foreground observations and a smaller but farther peak corresponding to the static background.
        The original CFTA formulation selects the dominant peak, whereas the proposed two-stage peak analysis identifies the farther background peak and determines the correct background threshold.
        }
    \label{fig:cfta_fine}
\end{figure}

Static background structures typically generate one or multiple dominant peaks in the empirical range histogram of a beam.
However, the composition of these histograms strongly depends on sensor characteristics, scene layout, and traffic density.
In high-traffic scenarios, foreground objects may form dominant peaks that can be mistakenly interpreted as background.

To address this issue, beams are categorized by their fraction of valid range observations during training.
Depending on this ratio $\rho_b$, three cases are distinguished:
\begin{itemize}
    \item \textbf{Noise case}: $\rho_b < \rho_{\text{noise}}$. The beam is considered unreliable and excluded from background modeling.
    \item \textbf{Sparse case}: $\rho_{\text{noise}} \le \rho_b < \rho_{\text{dense}}$. The beam contains a mixture of sparse background and dynamic objects.
    \item \textbf{Dense case}: $\rho_b \ge \rho_{\text{dense}}$. The beam provides sufficient support for stable peak estimation.
\end{itemize}
For sparse beams, a coarse histogram is evaluated for the presence of a dominant peak accounting for at least a fraction $\rho_{\text{static}}$ of all valid measurements.
If such a peak exists, the beam is processed using the dense-case procedure; otherwise, no background threshold is estimated for that beam (foreground-only).

For dense beams, we extend the original CFTA approach \cite{zhang_roadsidelidarvehicle_2022}.
CFTA assumes that the dominant histogram peak corresponds to the static background, which may not hold in real-world roadside LiDAR data.
Heavy traffic, occlusions, vegetation, thin structures, and low-reflectivity surfaces can produce competing peaks that violate the dominance assumption, as illustrated in Figure \ref{fig:cfta_fine}, where the dominant peak is not associated with the static background.
To combine robustness and precision, our threshold estimation uses a two-stage peak analysis: the coarse stage identifies a robust far-range candidate despite competing peaks, while the fine stage refines the threshold locally, avoiding the unreliable peak fragmentation that can occur when fine binning is applied over the full range.

\paragraph{Coarse Stage}
A coarse histogram is computed over the full range distribution.
We select $N_c$ bins with the highest counts and retain only those exceeding a minimum fraction of the total beam measurements.
Among the remaining candidates, the farthest peak is chosen.
A beam-specific trimming threshold is defined by adding a fixed offset to the selected peak location, and measurements beyond this limit are discarded.

\paragraph{Fine Stage}
Within the trimmed range, a refined histogram is constructed to obtain a higher-resolution estimate of the background peak.
As in the sparse case, we first test for a sufficiently dominant peak.
If none is found, we apply the same candidate selection strategy as in the coarse stage and select the farthest valid peak among the top $N_f$ bins.
Finally, the background threshold is obtained from the histogram bin with maximum distance to the line connecting the first bin and the selected peak, following the CFTA reference-line formulation proposed by Zhang et al. \cite{zhang_roadsidelidarvehicle_2022}.

\subsubsection{Quantile-Constrained Gaussian}

Histogram-based thresholding provides high modeling flexibility but requires careful peak selection and tuning.
As a simpler parametric alternative, we approximate the beam-wise background distribution with a quantile-constrained Gaussian model.

Beams with insufficient valid measurements are excluded from background modeling and are subsequently classified entirely as foreground.
For the remaining beams, we identify an upper quantile $q_{\text{hi}}$ of the range distribution and restrict the fit to samples above this quantile.
If too few samples are available, the interval is adaptively expanded.

A Gaussian distribution is then fitted to the selected samples using robust estimators.
The location parameter is given by the median,
\[
\mu_b = \operatorname{median}(r),
\]
and the scale is estimated via the Median Absolute Deviation (MAD) \cite{rousseeuw_alternativesmedianabsolute_1993},
\[
\sigma_b = 1.4826 \cdot \operatorname{median}(|r - \mu_b|).
\]

During inference, a measurement $r$ is classified as background if
\[
\frac{|r - \mu_b|}{\sigma_b} \le z_{\text{gate}},
\]
where $z_{\text{gate}}$ is a configurable gating parameter.

This approach replaces explicit histogram-based thresholds with a continuous, robust distance criterion and provides a compact unimodal approximation of the background distribution.

\subsubsection{Beam-Wise Gaussian Mixture}

\begin{figure}[t]
    \centering
    \def\svgwidth{0.98\linewidth}
    \input{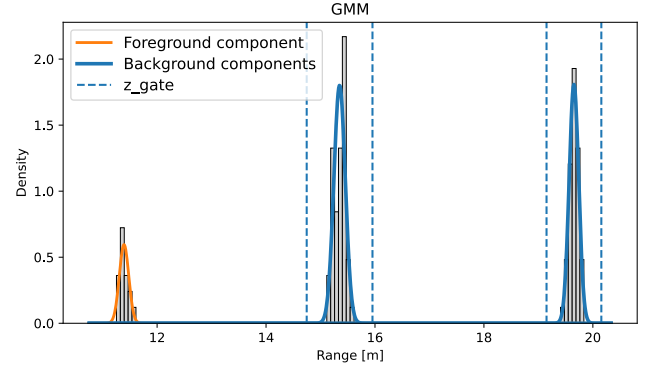}
    \caption{
        Example of a beam-wise range distribution modeled by a Gaussian mixture.
        Two components correspond to static background structures at different distances,
        while an additional component represents dynamic foreground observations.
        During inference, a measurement is classified as background if it lies within a gated distance $z_{\text{gate}}$
        of at least one component previously identified as background.
    }
    \label{fig:gmm_components}
\end{figure}

A single Gaussian or threshold may be insufficient when multiple stable structures are present along a beam.
Vegetation, thin objects, partial occlusions, or measurement variability can lead to multi-modal range distributions, as illustrated in Figure \ref{fig:gmm_components}.

To account for such effects, we model the range distribution of each beam using a one-dimensional Gaussian Mixture Model (GMM):
\[
p(r) = \sum_{k=1}^{K} w_k \, \mathcal{N}(r \mid \mu_k, \sigma_k^2).
\]

Beams with insufficient samples are excluded from modeling.
For the remaining beams, the number of components $K \in \{1,2,3\}$ is selected using the Bayesian Information Criterion (BIC), and parameters are estimated using the Expectation-Maximization (EM) algorithm \cite{dempster_maximumlikelihoodincomplete_1977}.
In contrast to unimodal methods, the mixture models the entire range distribution, including both foreground and background modes.

After training, components are classified as background based on stability constraints.
A component is considered background if it explains a sufficiently large fraction of observations and exhibits low dispersion,
\[
w_k \ge w_{\text{min}}
\quad \text{and} \quad
\sigma_k \le \sigma_{\text{max}}.
\]

During inference, a measurement $r$ is labeled as background if it lies sufficiently close to at least one component previously classified as background:
\[
\exists k \in \mathcal{B} :
\frac{|r - \mu_k|}{\sigma_k} \le z_{\text{gate}},
\]
where $\mathcal{B}$ denotes the set of components satisfying the stability constraints.

This formulation enables a multi-modal yet geometrically interpretable background representation in range space.
The increased modeling flexibility comes at the expense of higher computational cost compared to unimodal approaches.

\subsubsection{Image-Based Gaussian Mixture}

In addition to beam-wise modeling, we also evaluate a two-dimensional background subtraction strategy in the angular domain.
The LiDAR sampling structure naturally defines a 2D grid over elevation channels and azimuth angles.
For each frame, the minimum range per grid cell is stored and normalized to an 8-bit depth image.
This image representation is processed using the adaptive per-pixel Gaussian Mixture Model of Zivkovic \cite{zivkovic_efficientadaptivedensity_2006}, as implemented in OpenCV (MOG2).

Since the angular grid corresponds directly to beam indices, foreground/background decisions can be mapped back to the original 3D points without additional transformations.

To ensure comparability with the temporally stable background assumption, the learning rate during inference is fixed to zero, preventing further model adaptation after training.

\subsection{Spatial Consistency Filtering}

The background models described so far operate independently per beam for computational efficiency.
However, physical objects form spatially coherent structures in 3D space.
We therefore apply an additional spatial consistency filter to remove isolated residual foreground points.

\begin{table*}[t]
\caption{
Quantitative performance of geometry-based background subtraction methods on the HighwayScene dataset across three LiDAR technologies.
Results are reported without and with spatial consistency filtering.
Background models are optimized for recall-weighted $F_2$, while filtering parameters are optimized for $F_1$.
Precision (P), Recall (R), and corresponding $F_2$/$F_1$ scores are given in \%.
}
\label{tab:highwayscene_all_sensors}
\centering
\setlength{\tabcolsep}{5pt}
\begin{tabular}{llcccccc}
\toprule
\textbf{Sensor} & \textbf{Method}& \multicolumn{3}{c}{\textbf{No Consistency Filtering}}& \multicolumn{3}{c}{\textbf{With Consistency Filtering}} \\
\cmidrule(lr){3-5} \cmidrule(lr){6-8}
& & P & R & F2 & P & R & F1 \\
\midrule
\multirow{5}{*}{Ouster}
& CFTA \cite{zhang_roadsidelidarvehicle_2022} & 91.68 & 96.07 & 95.15 & 91.95 & 96.01 & 93.94 \\
& Density-Aware CFTA & 92.48 & 99.09 & 97.69 & 97.75 & 98.69 & 98.22 \\
& Quantile-Constrained GM & 91.39 & 99.54 & \underline{97.79} & 99.59 & 98.92 & \underline{99.25} \\
& Beam-Wise GMM & 94.00 & 99.74 & \textbf{98.53} & 99.47 & 99.39 & \textbf{99.43} \\
& Image-Based GMM \cite{zivkovic_efficientadaptivedensity_2006} & 89.54 & 95.40 & 94.17 & 99.51 & 94.55 & 96.97 \\
\midrule
\multirow{5}{*}{Blickfeld}
& CFTA \cite{zhang_roadsidelidarvehicle_2022} & 87.41 & 76.02 & 78.05 & 88.60 & 75.99 & 81.81 \\
& Density-Aware CFTA & 95.65 & 90.80 & 91.73 & 98.57 & 90.60 & 94.42 \\
& Quantile-Constrained GM & 88.79 & 93.73 & \underline{92.70} & 99.03 & 93.11 & \underline{95.98} \\
& Beam-Wise GMM & 93.37 & 95.50 & \textbf{95.06} & 98.71 & 95.25 & \textbf{96.95} \\
& Image-Based GMM \cite{zivkovic_efficientadaptivedensity_2006} & 83.43 & 89.72 & 88.39 & 98.95 & 88.92 & 93.67 \\
\midrule
\multirow{5}{*}{AEVA}
& CFTA \cite{zhang_roadsidelidarvehicle_2022} & 70.24 & 95.65 & 89.20 & 71.07 & 95.62 & 81.54 \\
& Density-Aware CFTA & 75.76 & 99.34 & \textbf{93.52} & 97.08 & 97.65 & \underline{97.36} \\
& Quantile-Constrained GM & 66.22 & 94.67 & 87.18 & 96.94 & 91.99 & 94.40 \\
& Beam-Wise GMM & 76.59 & 98.73 & \underline{93.33} & 98.56 & 96.64 & \textbf{97.59} \\
& Image-Based GMM \cite{zivkovic_efficientadaptivedensity_2006} & 62.66 & 96.57 & 87.14 & 97.30 & 92.50 & 94.84 \\
\bottomrule
\end{tabular}
\end{table*}

\subsubsection{Range-Adaptive Radius}

A common consistency check tests whether each foreground point has at least $n_{\text{min}}$ neighboring foreground points within a radius $r$.

Because LiDAR point density decreases with range, a fixed radius is inappropriate.
We therefore adapt the search radius as a function of range using the sensor's angular resolution:
\[
r_{\text{adaptive}}(d) = s \cdot d \cdot \tan(\Delta\theta_{\text{eff}}),
\]
where $d$ denotes the point range, $\Delta\theta_{\text{eff}}$ the effective angular spacing between beams, and $s$ a scaling factor. The resulting radius is computed individually for each foreground point.



\subsubsection{Local Angular Support}

Spatial consistency can also be enforced directly in the angular grid.
Instead of 3D neighbor searches, we apply a sliding window in the 2D beam grid and retain a foreground label only if at least $n_{\text{min}}$ neighboring cells provide support.

We consider two definitions of angular support:
\begin{itemize}
    \item \textbf{Range similarity}: Neighboring cells must exhibit similar ranges within a range-dependent tolerance $\tau(d)$.
    \item \textbf{Model consistency}: Neighboring cells support a foreground label if their underlying model decision is also foreground.
\end{itemize}

Cells without sufficient angular support are reclassified as background.
\section{Experiments}
\label{sec:experiments}

To ensure consistent and reproducible evaluation, we employ a strict train/validation/test protocol for both datasets.
Models are trained exclusively on unlabeled training data, parameters are selected on the validation split, and all results reported in Tables \ref{tab:highwayscene_all_sensors}--\ref{tab:coopscenes_results} are computed on the test split.

HighwayScene is split into 4{,}000 training frames, 1{,}000 validation frames, and 997 test frames.
For CoopScenes, labeled frames are split per scene, with roughly 50\% used for validation and the remainder for testing, while model training relies exclusively on unlabeled sequences without overlap with annotated maneuvers.

All methods operate on range-limited point clouds to standardize sensing volume across sensors and scenes, using 100\,m for HighwayScene and 150\,m for CoopScenes.
We use the published CFTA implementation unchanged, including its built-in consistency filter. 
All remaining methods use the same horizontal and vertical beam discretization and are paired with the proposed consistency filters.

Unless stated otherwise, all proposed models are trained on 600 frames (approximately one minute of recording).
More training frames did not yield further performance gains, indicating convergence of the background models.
Moreover, fixing the training horizon also reduces sensitivity to incidental foreground-sparse intervals and improves reproducibility.

For OpenCV MOG2, the inference learning rate is fixed to zero to prevent online adaptation and match the static-background assumption of the beam-wise models.

Evaluation is point-wise and aggregated over all test frames, with metrics computed for dynamic foreground points. Tuned parameters include the CFTA density and peak thresholds, the Quantile-Constrained Gaussian fit quantile and gate, the Beam-Wise GMM stability and gate thresholds, the MOG2 history and variance threshold, and the spatial-filter neighborhood and support thresholds; complete sweep ranges and selected values are reported in the repository.

All methods are evaluated in two stages. 
First, background model parameters are optimized for the recall-weighted $F_2$ score,
\[
F_\beta = (1+\beta^2)\frac{PR}{\beta^2 P + R}, \quad \beta = 2,
\]
which weights recall four times stronger than precision.
Second, spatial consistency filtering parameters are optimized for $F_1$.
This separation is motivated by the fact that the filtering stage only removes foreground predictions; consequently, recall cannot increase after background modeling and is therefore prioritized in the first stage. 

All experiments are conducted on an AMD 9950X CPU using multithreaded execution.

\begin{table}[b]
\caption{
Spatial consistency filtering strategies on HighwayScene (AEVA) using the beam-wise GMM. 
Precision (P), Recall (R), and $F_1$ scores are reported in \% and runtime (RT) in ms.
}
\label{tab:a42_filtering}
\centering
\setlength{\tabcolsep}{6pt}
\begin{tabular}{lcccc}
\toprule
\textbf{Consistency Filter} & P & R & F1 & RT \\
\midrule
None & 76.59 & 98.73 & 86.26 & -- \\

Radius (per-point) & 98.56 & 96.64 & \textbf{97.59} & \textbf{5.95} \\

Grid (range) & 97.34 & 97.14 & \underline{97.24} & \underline{22.41} \\

Grid (model) & 94.03 & 92.70 & 93.36 & 38.74 \\

\bottomrule
\end{tabular}
\end{table}

\subsection{HighwayScene}

\begin{table*}[t]
\caption{
Background subtraction performance on four CoopScenes scenes with varying foreground ratios (FG). 
Results are reported per scene using Precision (P), Recall (R), and $F_1$ score in \%.
}
\label{tab:coopscenes_results}
\centering
\setlength{\tabcolsep}{5pt}
\begin{tabular}{lcccccccccccc}
\toprule
& \multicolumn{12}{c}{\textbf{Scenes}} \\
\cmidrule(lr){2-13}
\textbf{Method}& \multicolumn{3}{c}{\textbf{Scene 1 (FG=3.14\%)}}& \multicolumn{3}{c}{\textbf{Scene 2 (FG=0.78\%)}}& \multicolumn{3}{c}{\textbf{Scene 3 (FG=0.59\%)}}& \multicolumn{3}{c}{\textbf{Scene 4 (FG=1.53\%)}} \\
\cmidrule(lr){2-4} \cmidrule(lr){5-7} \cmidrule(lr){8-10} \cmidrule(lr){11-13}
& P & R & F1 & P & R & F1 & P & R & F1 & P & R & F1 \\
\midrule
CFTA \cite{zhang_roadsidelidarvehicle_2022} & 65.84 & 94.76 & 77.70 & 50.41 & 97.62 & 66.48 & 48.52 & 95.59 & 64.37 & 50.20 & 93.61 & 65.35 \\

Density-Aware CFTA & 94.84 & 92.81 & \underline{93.82} & 92.69 & 93.55 & \textbf{93.12} & 92.62 & 90.42 & \textbf{91.51} & 99.61 & 93.78 & \underline{96.61} \\

Quantile-Constrained GM & 96.72 & 88.41 & 92.38 & 91.73 & 91.10 & 91.42 & 95.35 & 82.46 & 88.44 & 98.76 & 91.35 & 94.91 \\

Beam-Wise GMM & 97.78 & 77.93 & 86.73 & 92.76 & 91.20 & 91.97 & 90.72 & 81.47 & 85.85 & 98.41 & 95.09 & \textbf{96.72} \\

Image-Based GMM \cite{zivkovic_efficientadaptivedensity_2006} & 96.20 & 91.78 & \textbf{93.94} & 98.79 & 87.37 & \underline{92.73} & 95.75 & 85.24 & \underline{90.19} & 98.44 & 93.17 & 95.73 \\

\bottomrule
\end{tabular}
\end{table*}

Results on HighwayScene are summarized in Table \ref{tab:highwayscene_all_sensors} for three LiDAR technologies with substantially different scan patterns.
The beam-wise GMM achieves the best performance across sensors, consistent with its ability to represent multiple stable background modes that arise from vegetation, guard rails, and partial occlusions in roadside environments.

Notably, the unimodal Quantile-Constrained Gaussian and the Density-Aware CFTA remain competitive with the mixture model.
After applying spatial consistency filtering, the $F_1$ gap to the beam-wise GMM ranges from 0.18\% to 3.19\%, indicating that many beams in structured highway scenes can be approximated well by a single background mode.

In contrast, the original CFTA baseline delivers lower and more sensor-dependent performance. 
This is influenced by the construction of the angular grid used for histogram estimation. 
Depending on the available sensor metadata, the discretization may not fully match the true beam layout, particularly for non-uniform elevation patterns. 
Our implementation derives beam indices directly from the available scan pattern information, leading to more consistent behavior across sensors. 
The density-aware extension addresses peak robustness and is independent of scan pattern handling.

Spatial consistency filtering yields a large and consistent precision improvement across methods, particularly for the AEVA  where raw background modeling exhibits the strongest precision degradation.
With the proposed filtering enabled, the lowest precision across all methods and sensors exceeds 96\%.
For AEVA, precision increases by up to 35 percentage points at the cost of only a minor recall reduction, confirming that spatial consistency is essential for suppressing isolated false positives from noisy beam-wise decisions.
This result reflects the geometry-only scope of the benchmark: while the AEVA FMCW LiDAR additionally provides radial point velocities, using this information would make foreground separation in this highway scenario largely reducible to velocity-based filtering.
The velocity channel is therefore excluded from the evaluated methods to preserve a common geometric input representation across all sensors.

\subsection{Consistency Filtering}

Table \ref{tab:a42_filtering} compares the evaluated spatial consistency strategies. 
Both the per-point radius filter and the grid-based range filter achieve consistently strong and comparable $F_1$ scores, while the grid-based model-consistency filter performs substantially worse.

Across all experiments, the grid-based range filter yields the best result in 13 cases and the per-point radius filter in 15 cases, indicating no clear dominance of either approach. 
In the representative A42 example, the per-point radius filter improves $F_1$ from 86.26\% to 97.59\% at 5.95\,ms per frame, providing an attractive accuracy–efficiency trade-off.

The grid-based range filter has a runtime of 22.41\,ms in this configuration, reflecting the comparatively large angular neighborhood (2 vertical and 24 horizontal neighbors) used in the evaluation. 
Runtime scales predictably with the selected neighborhood size and remains compatible with real-time operation. 
In practice, the neighborhood is adapted to the specific sensor characteristics and scene requirements.

The grid-based model-consistency filter is both slower and less accurate. 
Relying on neighboring model decisions as support can propagate local modeling errors—particularly near sharp structural transitions such as guard rails or traffic signs—while increasing computational overhead.

\subsection{CoopScenes}

Table \ref{tab:coopscenes_results} reports results for four CoopScenes scenes with varying foreground ratios. 
Our density-aware CFTA achieves the most consistent performance across all scenes, while the original CFTA baseline remains clearly precision-limited.

Across scenes, performance remains high and no clear trend with respect to scene complexity or foreground ratio is observable. 
A notable exception is Scene 3, where temporal background stability is partially violated: a vehicle present during training is absent in validation and test sequences, leading to shadowed regions that are not correctly modeled as background.
This highlights a limitation of purely temporal background modeling under non-stationary static objects.

Across datasets and sensors, at least one proposed method achieves $F_1 > 90\,\%$ in every evaluated scene, demonstrating robust cross-sensor and cross-scene applicability.
Overall, beam-wise statistical modeling combined with spatial consistency filtering provides a practical and transferable solution for static roadside LiDAR background subtraction.

The qualitative example in Figure \ref{fig:intro_result} shows that the remaining errors after filtering are sparse and isolated.
\section{Conclusion and Future Work}
\label{sec:conclusion}

In this work, we introduced a systematic study of beam-wise statistical background modeling for statically mounted infrastructure LiDAR sensors.
Addressing the lack of reproducible and cross-sensor, multi-scene evaluations in roadside LiDAR background subtraction, we introduced the HighwayScene dataset, extended CoopScenes with static/dynamic annotations, and evaluated multiple complementary modeling paradigms under heterogeneous sensing conditions.

Across datasets, sensor technologies, and scene complexities, beam-wise statistical modeling proved to be a robust and transferable approach.
In particular, the density-aware CFTA achieved the most consistent overall performance across sensors and scenes, demonstrating strong robustness under varying traffic densities and background structures.
While the beam-wise Gaussian mixture model yielded the highest scores in structured highway scenarios, its performance was less stable in more complex urban environments.
At the same time, unimodal approaches such as the quantile-constrained Gaussian remained competitive in structured highway environments, indicating that many beams can be approximated well by a single dominant background mode.

A key finding of this study is the importance of spatial consistency filtering.
While beam-wise modeling provides high recall, spatial filtering substantially improves precision and stabilizes performance across sensors with different scan patterns and resolutions.
Combining lightweight per-beam statistical models with geometrically motivated filtering yields an efficient, real-time capable solution for infrastructure-based LiDAR background subtraction, largely independent of sensor resolution and range.

Limitations arise in scenarios where the assumption of temporally stable background is violated, which is unavoidable especially in urban scenes with parked vehicles.
Furthermore, the dependency on a known and fixed scan pattern remains a structural limitation of beam-wise formulations. 
Sensors with dynamically changing or insufficiently defined scan patterns remain challenging to model within a purely beam-wise background representation.
Future work will therefore investigate adaptive background updating strategies, for example by employing situationally adapted time windows for online training. In addition, more sensor-agnostic representations will be explored, including improved scan-pattern estimation without prior knowledge as well as alternative 3D voxel-based formulations.



\end{document}